\documentclass[runningheads]{llncs}

\usepackage{subfiles}
\usepackage[T1]{fontenc}
\usepackage{graphicx}
\usepackage{eccvabbrv}
\usepackage{multirow}
\usepackage{array}
\usepackage{subcaption}
\usepackage[export]{adjustbox}
\usepackage{graphicx}
\usepackage{booktabs}
\usepackage{amsmath}
\usepackage{amssymb}
\usepackage{cite}
\usepackage{hyperref}
\usepackage{cleveref}

\makeatletter
\newcommand{\@chapapp}{\relax}%
\makeatother
\usepackage{appendix}
\usepackage{lscape}
\usepackage{color}

\begin{document}

\title{Bounding Boxes and Probabilistic Graphical Models: Video Anomaly Detection Simplified}

\titlerunning{Video Anomaly Detection Simplified}

\author{
    Mia~Siemon\inst{1,2} 
    \and Thomas~B.~Moeslund\inst{2} 
    \and Barry~Norton\inst{1} 
    \and Kamal~Nasrollahi\inst{1,2} 
}
\authorrunning{M.~Siemon et al.}

\institute{Milestone Systems A/S, Banemarksvej 50, 2605 Brøndby, Denmark
\email{\{misi,kna,bno\}@milestone.dk} \and
Aalborg University, Rendsburggade 14, 9000 Aalborg, Denmark \\
\email{tbm@create.aau.dk}}

\maketitle

\begin{abstract}
In this study, we formulate the task of Video Anomaly Detection as a probabilistic analysis of object bounding boxes. We hypothesize that the representation of objects via their bounding boxes only, can be sufficient to successfully identify anomalous events in a scene. The implied value of this approach is increased object anonymization, faster model training and fewer computational resources. This can particularly benefit applications within video surveillance running on edge devices such as cameras. We design our model based on human reasoning which lends itself to explaining model output in human-understandable terms. Meanwhile, the slowest model trains within less than $7$ seconds on a $11^{th}$ Generation Intel Core i9 Processor. While our approach constitutes a drastic reduction of problem feature space in comparison with prior art, we show that this does not result in a reduction in performance: the results we report are highly competitive on the benchmark datasets CUHK Avenue and ShanghaiTech, and significantly exceed on the latest State-of-the-Art results on StreetScene, which has so far proven to be the most challenging VAD dataset. We release our code to the community at: \url{https://github.com/milestonesys-research/VAD-with-PGMs/}.

\end{abstract}

\begin{figure}
    \begin{center}
    \includegraphics[width=.85\linewidth]{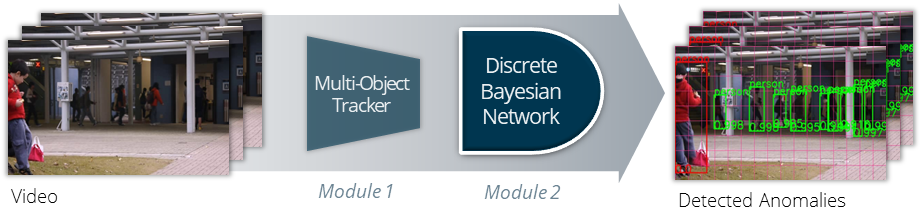}
    \end{center}
    \caption{Proposed Video Anomaly Detection Pipeline}
    \label{fig:pipeline}
\end{figure}

\section{Introduction}
\label{sec:intro}

The goal of Video Anomaly Detection (VAD)~\cite{IEEE-2020--Ramachandra-etal} solutions is to learn to differentiate between events which are commonly observed in a given scene, and those that are not. We follow accepted convention in referring to the former as \textit{normal} and the later as \textit{abnormal/anomalous}. Successful approaches in this domain of Computer Vision (CV) very often represent scene and object features in latent space to identify anomalous regions \cite{CVPR-2022--Ristea-etal,ICCV-2021--Liu-etal,CVPR-2023--Singh-Jones-LMiller}. In most of the cases, this entails the need for high-end accelerators during training and inference. Because of the evolving need for edge computing systems within the context of video surveillance~\cite{axis-2021--edge-computing} in industry, however, these solutions are less tractable compared to other approaches which are designed to run on low-power consumption devices~\cite{Sensors-2021--Cob-Parro-etal}. Inspired by this observation, the ambition of our proposed approach is to identify the smallest set of resources necessary to successfully perform VAD.

\medskip
Deep Learning (DL) has taken a powerful role in the areas of most prevalent CV tasks, and its capability to outperform humans in tasks like object detection is undeniable. Instead of solely relying on object detectors as done in prior art, however, our work explores the opportunity of exploiting models excelling in similar but more sophisticated CV tasks, such as Multi-Object Tracking (MOT)~\cite{AI-2021--Luo-etal,ECCV-2020--Wang-etal}. We show throughout the course of this paper that the output of such a MOT instance, the mere bounding box of a tracked object belonging to some class, combined with a Probabilistic Graphical Model (PGM)~\cite{book-2009--Koller,book-2019--Yang}, is sufficient to successfully perform VAD on currently most recognized benchmark datasets. Further, the exclusion of the deep features from latent space implies the restoration of a significant level of privacy in the process.

\medskip
Within the context of VAD, wherein we would like to determine what is normal and not, it appears intuitive to model observations as uncertainties within a framework which is based on conditional probability theory. The fundamental motivation for choosing a PGM in favor of a Conditional Variational Auto-Encoder, for example, as proposed by recent work of Liu~\etal~\cite{ICCV-2021--Liu-etal}, is that this particular Machine Learning (ML) concept enables the separation of knowledge and reasoning. Knowledge can be declaratively represented with clear semantics contained in graph models, moreover allowing for unambiguous explainability on model- and results-level.

\medskip
The contributions of this study are thus summarized as follows:

\begin{itemize}
    \item A new, simplified perspective on problem domain representation within VAD.
    \item A novel approach for fully object-centric VAD in which 8 high-level object bounding box attributes are learned by a discrete Bayesian Network to detect visual outliers in video streams.
    \item Human-understandable model and result explainability within the context of the high-level attributes used to model the problem domain.
    \item A new State-of-the-Art (SOTA) baseline for StreetScene significantly exceeding previous achievements by almost $4\%$ in terms of RBDC/TBDC average, and competitive RBDC/TBDC results on CUHK Avenue and ShanghaiTech.
\end{itemize}

\section{Related Work}
\subsubsection{Unsupervised Approaches}
Due to the unpredictable nature of anomalies, the task of VAD is mostly tackled by means of unsupervised learning techniques, and can be split into \textit{reconstruction}-based, \textit{distance}-based, and \textit{probabilistic} methods~\cite{IEEE-2020--Ramachandra-etal}. Most recent SOTA is primarily found in the first category. Very often, the given problem is split into proxy tasks, such that the most representative latent features are to be determined for an object's motion and its appearance by applying individually optimized Deep (Convolutional) Neural Networks (DNNs)~\cite{2021-JOBD--Alzubaidi-etal}. This can take place at the entire frame-level, or on local regions only. The most recent pioneering work includes: Georgescu \etal~\cite{CVPR-2021--Georgescu-etal,TPAMI-2021--Georgescu-etal}, Liu \etal~\cite{ICCV-2021--Liu-etal}, Ristea \etal~\cite{CVPR-2022--Ristea-etal}, and Doshi and Yilmaz~\cite{WACV-2023--Doshi-Yilmaz}.
Ramachandra and Jones~\cite{WACV-2020--Ramachandra-Jones} and Singh \etal~\cite{CVPR-2023--Singh-Jones-LMiller} on the other hand, proposed methods from the family of distance-based VAD approaches. They build their models using features defined in latent space and derive anomaly scores based on the nearest neighbor metric. These latent features represent spatio-temporal 3D volumes and capture normal observations around predefined regions that occur during training.

\subsubsection{Probabilistic Approaches}
In contrast to other approaches, probabilistic VAD methods approximate the training data by finding the most adequate probability distribution over the space of observations. This involves modeling a collection of different attributes by means of graph representation schemes, such as Bayesian or Markov Networks~\cite{book-2009--Koller, book-2019--Yang}, for example.
Spatio-temporal attributes are considered in Adam \etal~\cite{TPAMI-2008--Adam-etal}, Antić and Ommer~\cite{ICCV-2011--Antić-Ommer}, and Kim and Grauman~\cite{CVPR-2009--Kim-Grauman}. Saleh \etal~\cite{CVPR-2013--Saleh-etal} on the other hand focus on the detection of visual anomalies in single images, proposing an 'oddness' measure of their contents. The critical difference with respect to our approach is that the BN~\cite{CVPR-2013--Saleh-etal} operates on deep features generated by means of calibrated, discriminative attribute classifiers~\cite{CVPR-2009--Farhadi-etal}. Our network, on the other hand, operates on high-level object features and the temporal dimension, facilitating VAD across entire video streams. Another approach which follows a pipeline similar to that of~\cite{CVPR-2013--Saleh-etal}, was introduced by Ouyang and Sanchez~\cite{ICPR-2021--Ouyang-Sanchez}. Here, the authors propose the usage of Gaussian Mixture Models to conduct clustering of latent features which encode spatio-temporal information using two deep denoising Auto-Encoders.

\subsubsection{Explainability}
With the increasing emergence of ethical concerns related to Artificial Intelligence~\cite{eu-ai-act--2023}, there is a great focus on finding solution approaches to VAD which allow to also provide a certain level of explainability to their outcome, i.e., an answer to the question as to why a certain area/object has been flagged as anomalous by the system. Doshi and Yilmaz~\cite{WACV-2023--Doshi-Yilmaz} approached this challenge by including machine-generated human-readable scene understanding by means of Scene Graphs~\cite{TPAMI-2021--Chang-etal} into their pipeline. Similarly to Singh \etal~\cite{CVPR-2023--Singh-Jones-LMiller} we aim at providing a sufficient degree of explainability in terms of all spatio-temporal attributes we model. What separates our work is the intuitive and explainable modelling process of the problem domain by means of graph structures.

\section{Methodology}

Exploring the hypothesis that probabilistic reasoning over high-level attributes of object bounding boxes represents a sufficient mean to accurately detect anomalous events in video surveillance footage yielded the design of a pipeline which consists of two modules: one for image pre-processing, one to perform VAD. We deploy the MOT instance BoT-SORT~\cite{arXiv-2022--Aharon-etal} as the implementation of module one. This tracker assumes a closed set of object categories which may seem counter-intuitive in an open-world task such as VAD. We motivate this choice, however, by an observation made by Georgescu~\etal~\cite{TPAMI-2021--Georgescu-etal} who did not encounter "any significant false negatives due to the employment of a pre-trained object detector". We agree with the assumption that the majority of anomalies originate from humans and the behavior they exhibit. Therefore, we choose the object detector in our pipeline to be pretrained on the 'most common objects in context' defined in MS-COCO~\cite{ECCV-2014--Lin-etal}. 
Module two is a discrete Bayesian Network (BN), a particular type of PGM, which will be elaborated on in the remainder of this section. The output of our pipeline shown in~\Cref{fig:pipeline} is a probability score for each of the present bounding boxes, indicating how likely an object is to occur in the scene.

\subsection{Discrete Bayesian Network}
\label{subsec:discrete-BN}
The foundation of our study is a discrete BN which facilitates probabilistic reasoning over observations made in our VAD-specific problem domain. Each such observation is composed of a set of different variables that can independently influence the final outcome of this process. Embedded in \textbf{Bayes' Theorem}, each such variable is called \textit{random variable (RV)} and has a fixed set of possible values, known as its \textit{value space (VS)}. Together, all RVs span the sample space $\mathcal{S}$. The calculation of the resulting joint probability distribution to detect outliers is governed by domain-specific dependency structures. A well-informed choice of these variables and proper modeling of the dependency structures are crucial for any BN to be successfully conditioned on prior knowledge. Within VAD, such knowledge may depend on a variety of contexts relevant to dataset, scene, or maybe even recording perspective of the video footage. In this approach, we encode such information in our model via $8$ spatio-temporal variables which are derived from high-level appearance/motion attributes of a bounding box and its location within a frame.

\subsubsection{Random Variables for Anomaly Detection in Videos}
\label{subsection:random-variables}
In the description of all defined RVs and their corresponding VSs that follows in~\Cref{tab:random-variables} it can be seen that some VSs were defined in the numeric space while the rest is categorical. The motivation behind this modelling choice was purely to provide a high-level understanding of what the values of all these RVs statistically represent. For instance, we claim that it is much easier for a human to understand the conceptual difference between a \textit{small} and a \textit{large} bounding box. Even though the same applies to all object classes that are considered in this work, we refrain from listing all $80$ categories of the MS-COCO dataset in the list below and use their individual numerical identifiers instead. To further provide the BN with the capability of localizing anomalous events within a single frame, we divide the image into a uniform grid structure of quadratic cells like proposed by Siemon \etal~\cite{ICMV-2022--Siemon-etal}.

\begin{table}[htb]
  \caption{High-level Object Bounding Box Attributes Expressed as Spatial and Temporal Random Variables (RV). Counts $F_{\text{total}}$ and $G_{\text{total}}$ are dataset-dependent. Temporal RVs are marked with a single *.}
  \label{tab:random-variables}
  \centering
  \begin{tabular}{@{}l >{\centering\arraybackslash}m{1.0cm} l >{\centering\arraybackslash}m{1.1cm} l@{}}
    \toprule
    \textbf{Name} & \textbf{Node} & \textbf{Type} & \textbf{Count} & \textbf{Value Space (VS)} \\
    \midrule
    Frame & F & Numerical & $F_{\text{total}}$ & $\{f \, \in \, \mathbb{N}^+ \; | \; 1 \leq f \leq F_{\text{total}} \}$ \\
    Grid Cell & G & Numerical & $G_{\text{total}}$ & $\{g \, \in \, \mathbb{N}^+ \; | \; 1 \leq g \leq G_{\text{total}} \}$ \\
    Object Class & C & Numerical & 80 & $\{c \, \in \, \mathbb{N}^+ \; | \; 1 \leq c \leq 80\}$ \\
    
    Intersection Area & I & Categorical & $5$ & $\{$small, 1/4, 1/2, 3/4, full$\}$ \\
    \multirow{2}*{Box Size} & \multirow{2}*{BS} & \multirow{2}*{Categorical} & \multirow{2}*{$5$} & \multirow{2}{5.5cm}{$\{$x-small, small, medium, large, x-large$\}$}\\
    \\
    Box Aspect Ratio & BAR & Categorical & $3$ & $\{$portrait, landscape, square$\}$ \\
    
    \multirow{2}*{*Velocity} & \multirow{2}*{V} & \multirow{2}*{Categorical} & \multirow{2}*{$7$} & \multirow{2}{5.5cm}{$\{$idle, slow, normal, fast, very fast, super fast, lightning fast$\}$}\\
    \\
    *Direction & D & Categorical & $8$ & $\{$N, NE, E, SE, S, SW, W, NW$\}$ \\
    
  \bottomrule
  \end{tabular}
\end{table}

The thresholds for all categorical VSs are determined statistically per dataset by the standard deviation $\sigma$ and the mean $\mu$ of the underlying class-wise distribution of the given RV as observed during training. The data is discretized based on the observation's distance from the mean, expressed through multiples of $\sigma$. Regarding the temporal RVs defined in~\Cref{tab:random-variables}, it is left to mention that they are determined by the displacement between the two geometric center coordinates of the corresponding bounding boxes as the object is tracked across two consecutive frames. A visualization of the set of all spatial RVs is provided in~\Cref{fig:random-variables} for better understanding.

\subsubsection{Graphical Representation}
\begin{figure}[ht]
    \centering
    \begin{subfigure}{0.48\linewidth}
        \centering
        \includegraphics[width=0.8\linewidth, center]{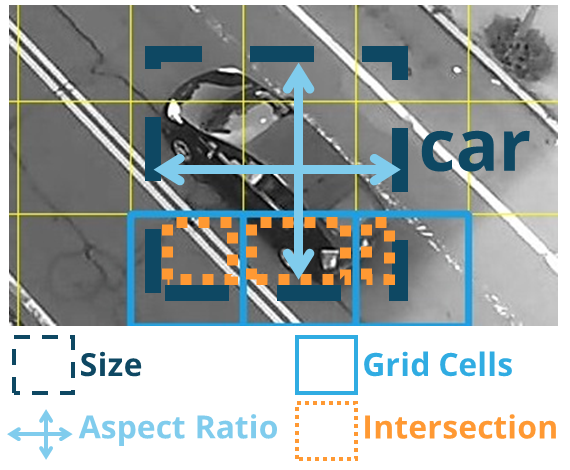}
        \caption{\textbf{Random Variables}}
        \label{fig:random-variables}
    \end{subfigure}
    \begin{subfigure}{0.48\linewidth}
        \centering
        \includegraphics[width=0.8\linewidth, center]{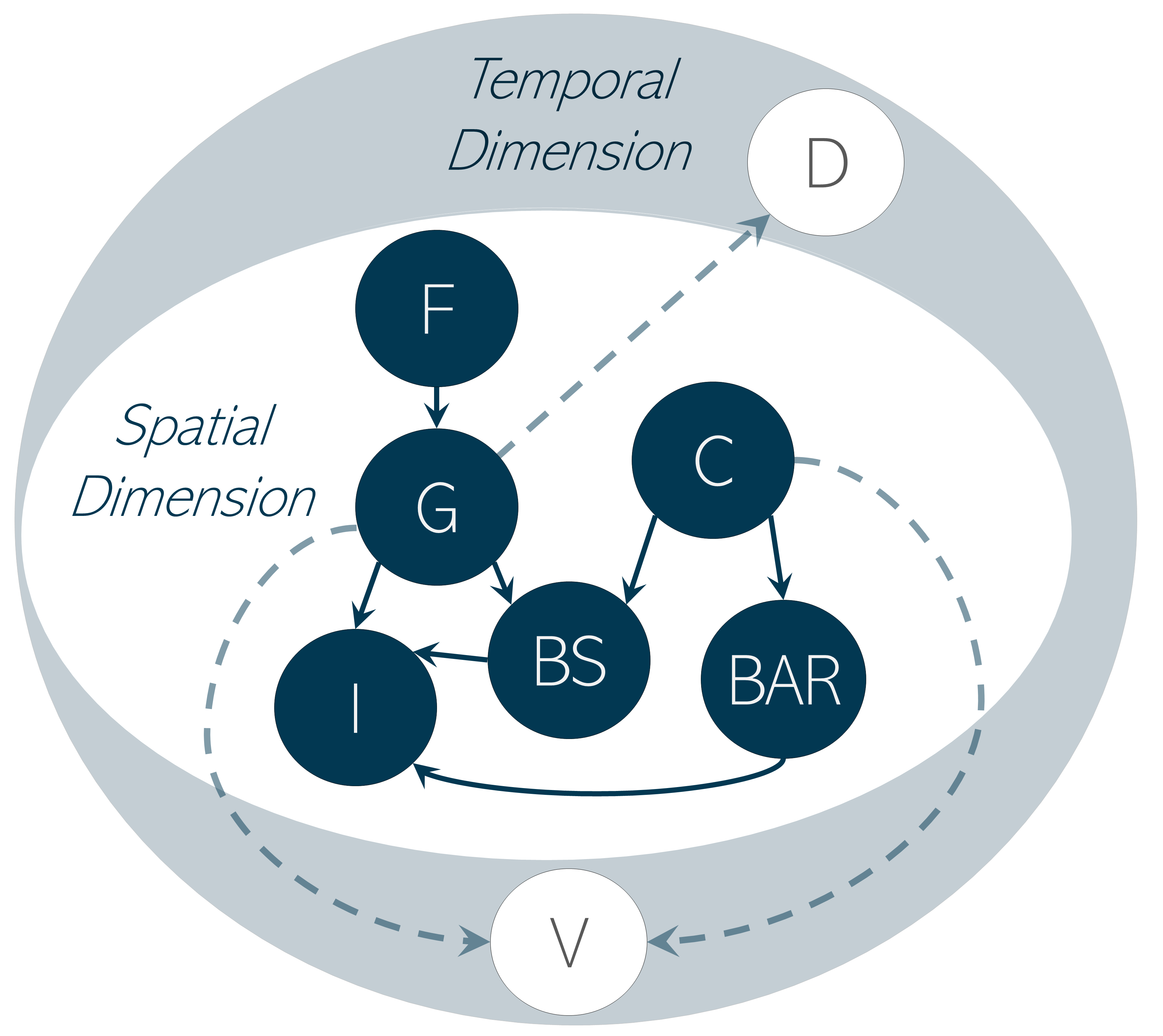}
        \caption{\textbf{Spatio-Temporal Model}}
        \label{fig:network-structure--spatio-temporal}
    \end{subfigure}
    \caption{\textit{Left:} All spatial RVs from~\Cref{tab:random-variables} except for the frame \textit{F} are illustrated on a sample image from StreetScene which was converted to greyscale for better visualization purposes. \textit{Right:} Our proposed BN model with conditional relations between all RVs to perform VAD.}
    \label{fig:network-structures}
    
\end{figure}

We now provide insight into how we express VAD-specific domain knowledge in terms of interdependencies between all RVs given in~\Cref{tab:random-variables}. This knowledge entails spatial and temporal clues to address appearance- and motion-based anomalies in videos. The resulting BN is given in the form of a Directed Acyclic Graph (DAGs) with unidirectional dependency connections between nodes (representative for RVs) such that two nodes are connected by at most one edge. Modelling the dependency structure as given by the directed edges in~\Cref{fig:network-structure--spatio-temporal} is motivated as follows: In order to give the problem domain we aim to represent a global context the frame node \textit{F} is serving the purpose of the root node in the graph. While each frame is divided into a uniform grid structure composed of grid cells G, the intersection \textit{I} between an object's bounding box and the respective grid cell \textit{G} is constrained by the aspect ratio \textit{BAR} and the size of the box \textit{BS}. Drawing a dependency between grid cell \textit{G} and the bounding box size \textit{BS} aims at encoding the scene depth given the influence it can have on the perceived size of some observed object. Following this modelling approach, both the size \textit{BS} and aspect ratio \textit{BAR} of a bounding box are constrained according to the class the observed object is representing. To illustrate this with an example, it can be claimed that the bounding box sizes of vehicles are on average larger than those drawn around humans, including the fact the the most common aspect ratio of a car resembles on average the landscape format. The bounding box of a human captured on CCTV footage in the city on the other hand would most likely be assumed to be of portrait format.

We express the temporal dimension of the model via the velocity \textit{V} and direction \textit{D} of the object. Following the notion of encoding scene depth in our model, the velocity \textit{V} of a bounding box is not only constrained on the object class (humans tend to move slower than cars, e.g.) but also on the area of the frame in which it is moving (planes flying in the sky appear to move relatively slow when observed from the ground, e.g.).

\subsubsection{Parameter Learning}
The term \textit{learning} in the context of PGMs denotes the process of deriving the optimal set of probability estimates for all possible events which are conditioned on certain prior observations. In other words, once the graphical representation of the problem distribution has been set, the goal of parameter learning is to provide the means to perform probabilistic reasoning. This can be achieved by constructing an approximation of the so-called \textit{joint probability distribution} of the given RVs and their VSs. Popular optimization algorithms for conducting parameter learning are: Maximum Likelihood Estimator (MLE), Bayesian Estimator and the Expectation-Maximization (EM) algorithm. While EM is primarily used in cases in which data is incomplete, the Bayesian approach is of advantage when only a limited amount of observations is available. Given the sizes of current benchmark VAD datasets, MLE was chosen for fitting the classifier to the training data. Briefly, the aim of Maximum Likelihood Estimation is to maximize the likelihood function describing our probabilistic model. Since this model is parameterized by a vector $\boldsymbol{\theta}$ containing the set of parameters of all RVs, the likelihood function is equivalent to the mean of how the obtained probabilities change with respect to different values of $\boldsymbol{\theta}$. In other words, the likelihood function estimates the probability, also called \textit{density}, assigned to the training data by the model given a particular choice of parameters.

\subsubsection{Inference}
During inference in PGMs the generated joint probability distribution is queried in order to obtain the posterior probabilities for events which occurred under the presence of prior observations, also known as \textit{evidence}. In discrete BNs this may be accomplished by Variable Elimination or Belief Propagation. Choosing the former for this study the actual detection of anomalies is performed by extracting a probability score from the discrete BN model for all objects which were detected in the test set. These scores indicate how likely a given object is to appear in the scene, with lower values pointing towards possible anomalies. Given that the class of the object is known upfront, all remaining evidence is gathered and supplied to the query which retrieves the Conditional Probability Table (CPT) for all classes at a certain grid cell. In mathematical terms, this results in the computation of $P(C \; | \; G, I, BS, BAR, V, D)$. Looking up the detected class in this CPT, the object's probability/anomaly score is averaged over every cell covered by the bottom edge of its bounding box. It is important to note that if a detected class does not exist in the CPT, a score of 0.0 is assigned. Similarly to~\cite{TPAMI-2021--Georgescu-etal}, we also employ a Gaussian filter to smoothen anomaly scores extracted on a frame-level.

\section{Experiments}
\subsection{Datasets}
\label{subsection:datasets}
Experiments were conducted on three currently popular publicly available VAD datasets: CUHK Avenue~\cite{ICCV-2013--Lu-etal}, ShanghaiTech~\cite{CVPR-2018--Liu-etal} and StreetScene. Even though our primary objective is to conduct VAD in single-scene scenarios due to the underlying grid structure in our approach, we also included the multi-scene dataset ShanghaiTech in our experiments. The poor frame coverage of objects in training videos of some of the scenes along with the argumentation given by Ramachandra and Jones~\cite{WACV-2020--Ramachandra-Jones}, however, led us to train a single model on all scenes of the dataset. This approach was also supported by the comparable variance of object appearances across the scenes, and the fact that the walking areas share approximately the same frame location between scenes.

\subsubsection{Frames}
Results presented in~\Cref{tab:quantitative-results--1} are based on experiments that involved either every third training frame (CUHK Avenue) or every fifth one (ShanghaiTech and StreetScene). Reducing the amount of training images was primarily motivated by statistical models being prone to over-fitting when faced with too much training data. Additionally, it naturally implies less processing time during both, training and inference. Both aspects are supported by corresponding data presented in the supplementary material.

\subsubsection{Grid Cells and Score Fusion}
In our approach, we reasoned over observations made at different grid cell granularities at a time. The individual resolutions (CUHK Avenue: $(640\times360)$, ShanghaiTech: $(1\,280\times720)$ and StreetScene: $(1\,280\times720)$) yielded these cell sizes to equal to $[20,40]$ for CUHK Avenue, and $[40,80]$ for the two latter, respectively. During inference, we fused the scores of the different granularities to then obtain the final object probabilities.

\subsection{Data Preprocessing}
We utilize BoT-SORT to pre-process all training and test frames. We choose it to be based on the object detector YOLOv7, pretrained on the MS-COCO dataset, including an object re-identification module that was trained on the Multiple Object Tracking 17 (MOT17) dataset~\cite{arXiv-2016--Milan-etal}. In order to alter the tracker's capabilities of catching very small objects in video frames we additionally finetune the object detector on the VisDrone-DET\cite{TPAMI-2021--Zhu-etal} dataset which belongs to the collection of datasets recorded out of Unmanned Aerial Vehicles (UAV) such as drones from extreme bird's eye view perspectives.

\subsubsection{Confidence Threshold}
Avoiding unnecessary noise in the discrete BN model, we filtered all objects coming from the MOT module according their corresponding confidence score given a certain threshold. This threshold was determined in accordance with the overall distribution of confidence scores of boxes detected during training on a particular scene and per dataset. It was set dynamically prior to the generation of observations: one for the class \textit{person}, and another one for all remaining classes. This differentiation was motivated by the fact that common object detectors are strongly biased towards detecting human beings. Various experiments showed that the confidence threshold yielding best results across all datasets is approximately $2\cdot\sigma$ smaller than its mean $\mu$.

\subsection{Training}
\label{subsec:training}
By using the term \textit{training}, we refer to the estimation of the joint probability distribution spanned by $\mathcal{S}$ described in~\Cref{tab:random-variables}. The data used for this purpose is fully discrete, and can therefore be encoded in a table of observations. Following the design of the network structure presented in~\Cref{fig:network-structure--spatio-temporal}, a single observation consists of a fixed set of features comprising assignments to the RVs of $\mathcal{S}$. For any grid cell intersecting with any bounding box in a given frame, one such observation is created and added to our tabular dataset. We process the bottom part of a bounding box which affects only a proper subset of cells in a particular row of the grid structure. This is motivated by the finding that a full-box approach can result in excessive noise when confronted with significant occlusions between objects. We supply corresponding proof in the supplementary material.

\subsection{Technical Details}
The core of this work is based on Python 3, PyTorch v1.11.0 and \texttt{pgmpy}, an open-source Python implementation of Probabilistic Graphical Models. The principal components of the hardware used for benchmarking are: an NVIDIA GPU, model GeForce RTX 3080 Ti, with 12GB of memory, running CUDA 11.3, and an 11th Gen Intel(R) Core(TM) i9-11900K @ 3.50GHz CPU.

\paragraph{Execution Time of VAD Module}
In its current implementation the training and prediction phase of our PGMs is conducted on the CPU mentioned above. Reaching a processing speed of almost up to 30 cells per second during inference while the training completes in less than a second (both are dataset dependent), we refer to the supplementary material for further details on this note.

\subsection{Evaluation Metrics}
\label{subsection:evaluation}
In choosing evaluation metrics, we follow the majority of published research on VAD: Frame-level Area Under the Curve (AUC), and Region-Based Detection Criterion (RBDC) and Track-Based Detection Criterion (TBDC). The latter two have been proposed by Ramachandra and Jones~\cite{WACV-2020--Ramachandra-Jones} in order to replace the previously used frame-based criterion because it is neither capable of locating an anomaly within a frame nor accounting for different amounts of anomalous regions. This metric is thus only roughly indicative of a method's true performance. Similarly to Singh \etal~\cite{CVPR-2023--Singh-Jones-LMiller}, we report frame-level AUC scores nevertheless, for completeness reasons. The calculation of both RBDC and TBDC scores has been facilitated by a script provided by Georgescu \etal alongside their work described in~\cite{TPAMI-2021--Georgescu-etal}. Given the large discrepancy between RBDC and TBDC scores and to thus allow for a simpler comparison of the said metrics, we add an additional column to~\Cref{tab:quantitative-results--1} containing their average value.

\subsection{Quantitative Results}

Results shown in~\Cref{tab:quantitative-results--1} indicate strong performance of our proposed approach across all three benchmark datasets especially given that it is solely based on the analysis of high-level bounding box features and their particular location with the frame. Our model achieves $66.14\%$ on RBDC/TDBC average which marks the third-best result to date on CUHK Avenue. Similarly, the results achieved on \textbf{ShanghaiTech} remain competitive. Scoring third-best in terms of RBDC scores reported to date on this dataset also makes this method a close runner-up on RBDC/TBDC average. Further, despite the level of difficulty introduced by \textbf{StreetScene}, we set a new SOTA baseline with our spatio-temporal discrete BN model, and report $30.65\%$ in RBDC and $66.03\%$ in TBDC, surpassing Singh \etal~\cite{CVPR-2023--Singh-Jones-LMiller} by nearly $4\%$ on RBDC/TBDC average.

\label{subsection:results}
\begin{table}[!hpb]
    \caption{\textbf{Accuracy Scores (\%)}: Frame-level AUC, RBDC, TBDC, and the mean $\mu_\text{R,T}$ of the two latter. Stressing \textbf{highest}, \underline{second-highest} and \textit{third-highest} results per metric.}
    \label{tab:quantitative-results--1}

    \bigskip
    
    \begin{subtable}[t]{1.0\textwidth}
        \centering
        \begin{tabular}{@{}|l|c|c|c||c|c|c|c||c|@{}}
            \hline
             & \multicolumn{4}{c|}{\textbf{CUHK Avenue}} & \multicolumn{4}{c|}{\textbf{ShanghaiTech}} \\
            Method & Frame & RBDC & TBDC & $\mu_\text{R,T}$ & Frame & RBDC & TBDC & $\mu_\text{R,T}$ \\
            \hline\hline
            Ramachandra \etal\cite{WACV-2020--Ramachandra-Jones} & 72.00 & 35.80 & 80.90 & 58.35 & 61.00 & 21.00 & 53.00 & 37.00 \\
            Georgescu \etal\cite{TPAMI-2021--Georgescu-etal} & \textit{92.30} & \textit{65.05} & 66.85 & 65.95 & \underline{82.70} & 41.34 & 78.79 & 60.07 \\
            Georgescu \etal\cite{CVPR-2021--Georgescu-etal} & 91.50 & 57.00 & 58.30 & 57.65 & \textit{82.40} & 42.80 & \textit{83.90}  & 63.35 \\
            Liu \etal\cite{ICCV-2021--Liu-etal} & 89.90 & 41.05 & \textit{86.18} & 63.62 & 74.20 & 44.41 & 83.86 & \textit{64.14} \\
            Ristea \etal\cite{CVPR-2022--Ristea-etal} +~\cite{ICCV-2021--Liu-etal} & 90.90 & 62.27 & \textbf{89.28} & \underline{75.78} & 75.50 & \underline{45.45} & \underline{84.50} & \underline{64.98} \\
            Ristea \etal\cite{CVPR-2022--Ristea-etal} +~\cite{TPAMI-2021--Georgescu-etal} & \textbf{92.90} & \underline{65.99} & 64.91 & 65.45 & \textbf{83.60} & 40.55 & 83.46 & 62.01 \\
            Singh \etal\cite{CVPR-2023--Singh-Jones-LMiller} & 86.02 & \textbf{68.20} & \underline{87.56} & \textbf{77.88} & 76.63 & \textbf{59.21} & \textbf{89.44} & \textbf{74.33} \\
            \hline
            Ours & \underline{92.72} & 60.18 & 72.09 & \textit{66.14} & 61.28 & \textit{45.40} & 81.87 & 63.64 \\
            \hline
        \end{tabular}
    \end{subtable}

    \medskip

    \begin{subtable}[b]{1.0\textwidth}
        \centering
        \begin{tabular}{@{}|l|c|c|c||c|c|c|c||c|@{}}
            \hline
             & \multicolumn{4}{c|}{\textbf{StreetScene}} \\
            Method & Frame & RBDC & TBDC & $\mu_\text{R,T}$ \\
            \hline\hline
            Ramachandra \etal\cite{WACV-2020--Ramachandra-Jones} & \underline{61.00} & \textit{21.00} & \textit{53.00} & \textit{37.00} \\
            Singh \etal\cite{CVPR-2023--Singh-Jones-LMiller} & - & \underline{24.30} & \underline{64.50} & \underline{44.40} \\
            \hline
            Ours & \textbf{72.70} & \textbf{30.65} & \textbf{66.03} & \textbf{48.34} \\
            \hline
        \end{tabular}
    \end{subtable}
    
\end{table}

\subsection{Qualitative Results: Explainability}
\label{subsec:explainability}
\begin{figure}[!ht]
    \centering
    \includegraphics[width=1.0\textwidth, center]{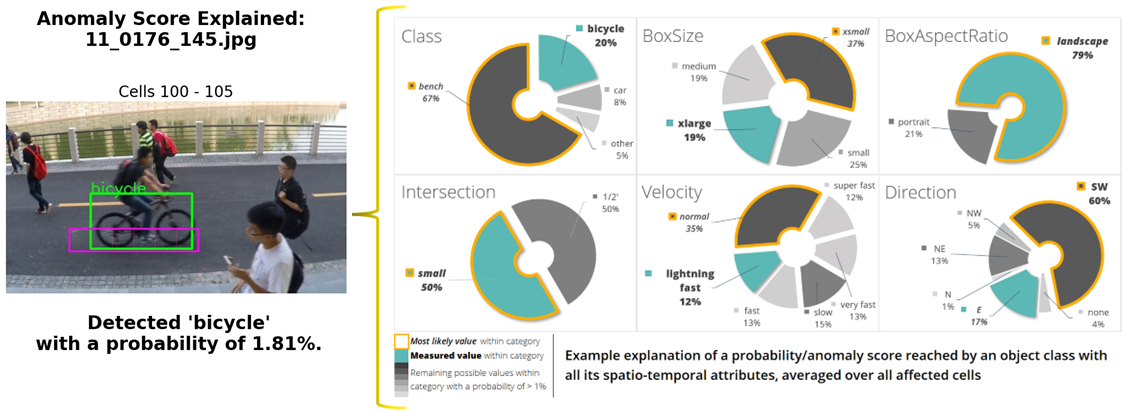}
    \caption{A visualization proposal of explaining the anomaly score extracted for a set of cells given the current appearance/velocity expressed of the bounding box through RVs defined in~\Cref{tab:random-variables}.}
    \label{fig:results-explained--shanghaitech}
\end{figure}

Given an object of very low probability, i.e., an anomaly, observed in some cells during test time, we aim at finding a statistical explanation for it. To this end, we present an example visualization of an anomaly taken from test video $11$ of the benchmark dataset ShanghaiTech in~\Cref{fig:results-explained--shanghaitech}. Based on the breakdown of individual probability distributions of all visual attributes we defined in~\Cref{tab:random-variables}, it can be deducted that the detected object ('bike') was flagged as anomalous in the given area due to the observed values of attributes \textit{Class} ('bicycle'), \textit{BoxSize} ('xlarge'), \textit{Velocity} ('lightning fast') and \textit{Direction} ('E'). More likely values for these categories based on the training data, highlighted with a thick orange border, would correspond to 'bench', 'xsmall', 'normal' and 'SW', respectively. In other words, the object is more likely to be a 'bench' under the remaining observed evidence. If it indeed is a 'bicycle', however, then the movement direction is more likely to be 'SW' given the remaining evidence, and so on. The charts show the average values of extracted probabilities per cell intersecting with the bottom border of the object bounding box. The total corresponding frame area is highlighted with a pink rectangle. We provide similar explainability visualizations for test samples from the remaining benchmark datasets in the supplementary material.

\subsection{Ablation Study}
\begin{figure}[ht]
    \centering
    \includegraphics[width=1.0\linewidth]{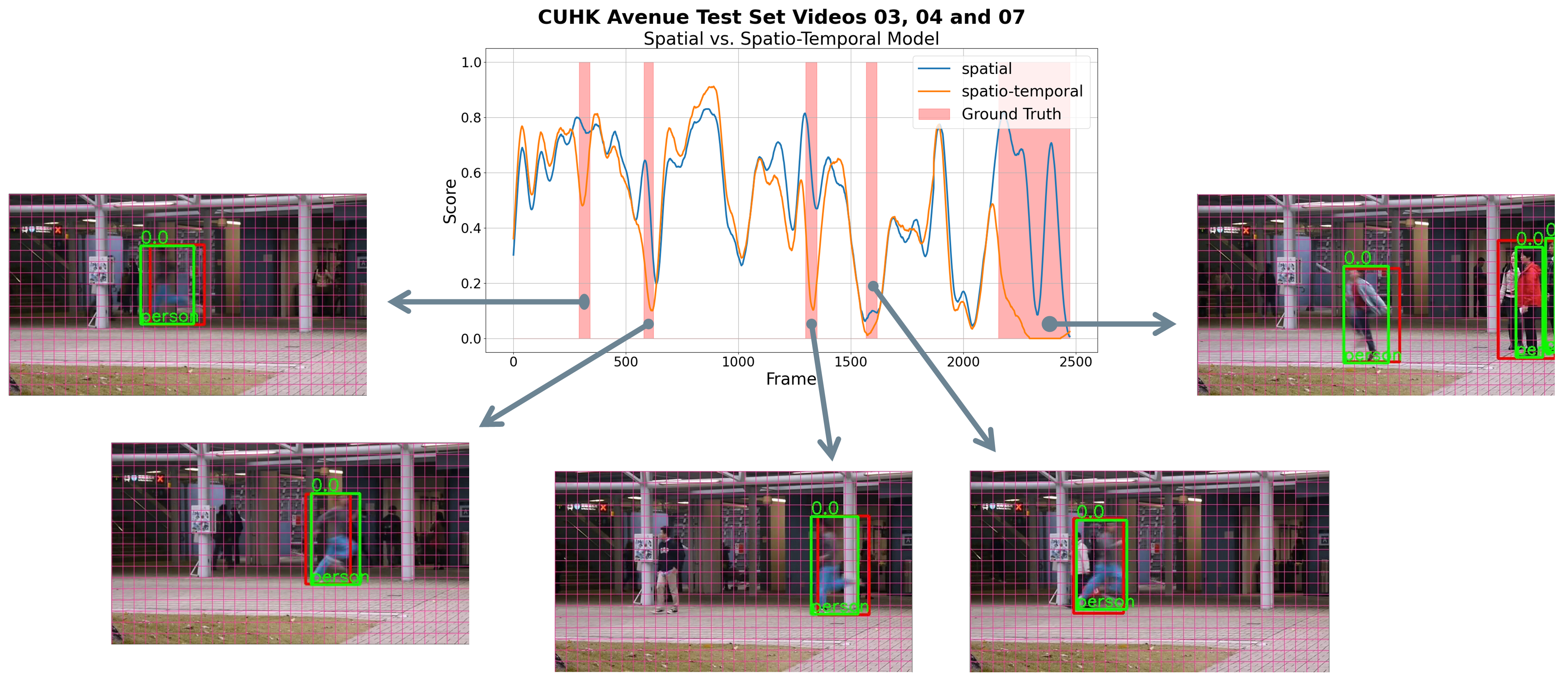}
    \caption{Contrasting spatial and spatio-temporal model versions~(\Cref{fig:network-structure--spatio-temporal}) based on three concatenated CUHK Avenue test videos, containing only temporal anomalies (a man running, a child jumping) in terms of frame-level AUC scores. The images are generated by the spatio-temporal model version.}
    
    \label{fig:Avenue--spatio-temporal--comparison}
    
\end{figure}

In the following, we want to emphasize the effect of the temporal dimension in our discrete BN shown in~\Cref{fig:network-structure--spatio-temporal} on the detection of motion-based anomalies in videos. To this end, we create a purely spatial counterpart by removing the temporal dimension of our spatio-temporal model only keeping the inner spatial dimension. To contrast the performances of both network versions, we generated~\Cref{fig:Avenue--spatio-temporal--comparison} based on three concatenated test videos. We chose test videos \#03, 04 and 07 of CUHK Avenue for this demonstration purpose because of the little background noise and the clearly visible temporal anomalies they contain. The enhanced capability of detecting temporal anomalies with our spatio-temporal model version is indicated here by the clear drop in frame-level AUC in areas highlighted with red background, i.e., anomalous frames.

\section{Conclusions and Future Work}
\label{sec:conclusions}
In conclusion, PGMs proved to be very suitable candidates to tackle VAD when combined with most recent MOT models. While many other works are challenged by StreetScene, we have shown that the modeling framework we used delivers new SOTA baselines on all recognized metrics on this dataset. We hypothesize that our model is not capable of beating previous SOTA results on CUHK Avenue and ShanghaiTech as opposed to StreetScene because of the anomaly types the former two datasets contain. Often, these anomalies depict anomalous human behavior such as throwing papers in CUHK Avenue, and dropping/playing with objects in ShanghaiTech. Unless such behavior results in a significant change of observed RV values, our method will not be capable of detecting such anomalies. Nevertheless, we believe that the results on all benchmark datasets prove that the representation of objects via their surrounding bounding boxes is indeed sufficient to successfully detect anomalies in videos, representing a novel and valuable insight into the domain of VAD. Another advantage of deployed ML models is that they are capable of performing better with less training data which we also conclude based on our experiments. Through these means, we foresee that relative advantages can be found in computational complexity, and both space and time demands.

\paragraph{Future Work} The first aspect we would like to address concerns the fact that the deployed PGM is operating on discretized variable values. While this can be embedded in a relatively stable statistical setup, it has limitations when compared to a purely continuous dimension. The exploration of continuous PGMs for VAD is thus left for future work. Second, given the rigid grid structure which is imposed on the video frames, it is debatable whether there might exist more efficient alternatives to induce localization capabilities to the model, and is therefore also left for future work. Third, as far as our experiments are concerned, the pipeline is only as good as the closed-set tracking module used, with its error rate being propagated to the PGM. We leave the validation of this hypothesis for future work, on the assumption that a single-scene VAD dataset with GT tracking labels for training comes to be available. This includes the implementation of open-world object tracking and/or detection modules.


\bibliographystyle{splncs04}
\bibliography{references}

\clearpage

\section{Supplementary Material}
\begin{appendices}
\label{sec:appendices}
\renewcommand{\thesection}{\appendixname~\Alph{section}}

In the following we present accompanying material for our work \textit{Bounding Boxes and Probabilistic Graphical Models: Video Anomaly Detection Simplified}. As indicated in the main paper, additional information and data is provided with respect to ablation studies concerning comparisons regarding the amount of processed training data, computational complexity imposed on the hardware, and more visualizations of qualitative results, each of which is addressed separately in the respective sections below.

\section{Computational Complexity}
\label{appendix:processing-time}
In this part of the appendix the focus is put on disclosing the computational complexity of our VAD module, i.e., the (average) execution times during the training and testing phase. This is illustrated based on the three datasets CUHK Avenue, StreetScene and ShanghaiTech. It is important to mention on this note that those values do not include the preparation time of the training data, i.e., the generation of training observations as described in Section~4 of the main paper. Results reported in~\Cref{tab:computational-complexity--bottom-border} entail thus the following information:

\subsubsection{Cell Size and Resolution} All experiments were conducted based on a smaller and a larger grid granularity. Sizes may vary across datasets given the corresponding resolution their camera footage comes in.

\subsubsection{Training Set} The reduction of training frames was performed by means of a \textit{slice factor}. Setting it equal to 5 and/or 3 results in the inclusion of every fifth and/or third frame into the observation generation process, while 1 means that the entire training set is considered. While the total number of objects is only dependent on the slice factor, and not on the cell size, the amount of generated observations equals to the overall number of intersections that were recorded between grid cells and the bottom edge of object bounding boxes.

\subsubsection{Training Time (sec)} This value represents the total execution time of the \texttt{pgmpy} library function responsible for calculating the joint probability distribution over all RVs.

\subsubsection{Average Inference Time (sec)} Here, we document the average inference time of each dataset on a cell-, object- and frame-level. Each process was running on a single CPU, specifically an 11th Gen Intel(R) Core(TM) i9-11900K @ 3.50GHz. We believe that a significant speed up can be achieved by means of appropriate parallelization mechanisms, which are missing in the current implementation and left for future work.

\section{Bounding Box - Grid Cell Intersections}
\begin{figure}[ht]
    \begin{subfigure}{0.48\linewidth}
        \includegraphics[width=1.0\linewidth, left]{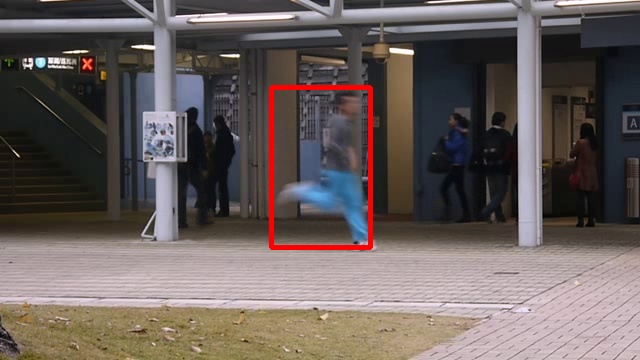}
        \caption{\textbf{Whole Box} - Frame-level AUC: $55.54$, RBDC: $28.12$, TBDC: $44.15$}
        \label{subfig:full-box}
    \end{subfigure}
    \begin{subfigure}{0.48\linewidth}
        \includegraphics[width=1.0\linewidth, right]{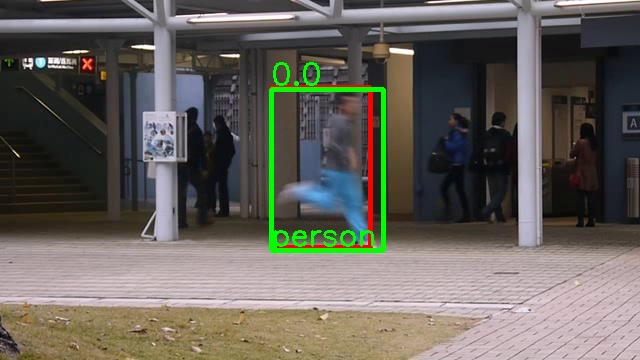}
        \caption{\textbf{Bottom Border} - Frame-level AUC: $74.00$, RBDC: $60.63$, TBDC: $98.59$}
        \label{subfig:bottom-border}
    \end{subfigure}
    \caption{Contrasting different observation generation techniques. Reported results (\%) are based on test video \#03 of CUHK Avenue. Ground Truth annotations are drawn in red, and detections in green (0 = anomalous, 1 = normal).}
    \label{fig:full-box--vs--bottom-border}
\end{figure}

In~\Cref{fig:full-box--vs--bottom-border}, we illustrate the effect achieved by only considering the intersections of cells with the bottom border of a bounding box, in contrast to recording all cells which are partially or fully covered by the object. This effect becomes especially significant when very far-reaching scenes are contained in a dataset, such as CUHK Avenue, in which significant occlusions between objects can result in excessive noise. \Cref{subfig:full-box} shows the result of training the PGM based on all cells which are affected by an object's bounding box. Clearly, this particular version of the model is not capable of detecting a human moving at an anomalous speed across the scene. \Cref{subfig:bottom-border}, on the other hand, shows how the noise in the training data is reduced when the processed samples are reduced to include cells intersecting with the bottom border of an object's bounding box only.

\begin{landscape}
    \begin{table}
    
        \caption{Computation Complexity during Training and Inference of our Spatio-Temporal Model}
        \label{tab:computational-complexity--bottom-border}
        \begin{center}
            \begin{tabular}{ m{2.0cm} |c||c|c|c|c||c|c|c|c||c|c|c|c}
            
            \multicolumn{2}{c||}{} & \multicolumn{4}{c||}{\textbf{CUHK Avenue}} & \multicolumn{4}{c||}{\textbf{StreetScene}} & \multicolumn{4}{c}{\textbf{ShanghaiTech}} \\
            \hline\hline
            \multicolumn{2}{c||}{Cell Size} & \multicolumn{2}{c|}{40} & \multicolumn{2}{|c||}{20} & \multicolumn{2}{c|}{80} & \multicolumn{2}{|c||}{40} & \multicolumn{2}{c|}{80} & \multicolumn{2}{|c}{40} \\
            \hline
            \multirow{2}{2.0cm}{Resolution} & pixels & \multicolumn{4}{c||}{$(640 \times 360)$} & \multicolumn{8}{c}{$(1~280 \times 720)$} \\
            \cline{2-14}
             & cells & \multicolumn{2}{c|}{$(16 \times 9)$} & \multicolumn{2}{|c||}{$(32 \times 18)$} & \multicolumn{2}{c|}{$(16 \times 9)$} & \multicolumn{2}{|c||}{$(32 \times 18)$} & \multicolumn{2}{c|}{$(16 \times 9)$} & \multicolumn{2}{|c}{$(32 \times 18)$} \\
            \hline\hline
            \multirow{4}{2.0cm}{Training Set} & frames & \multicolumn{4}{c||}{15~328} & \multicolumn{4}{c||}{56~846}  & \multicolumn{4}{c}{269~732}\\
            \cline{2-14}
             & slice factor & 3 & 1 & 3 & 1 & 5 & 1 & 5 & 1 & 5 & 1 & 5 & 1 \\
            \cline{2-14}
             & objects & 58~006 & 173~938 & 58~006 & 173~938 & 61~663 & 308~397 & 61~663 & 308~397 & 218~598 & 1.093M & 218~598 & 1.093M \\
            \cline{2-14}
             & observations & 100~640 & 301~835 & 148~838 & 446~477 & 143~206 & 716~054 & 223~479 & 1.117M & 351~310 & 1.757M & 488~322 & 2.442M \\
            \hline\hline
            Training Time (sec) & per dataset & 1.6108 & 1.6387 & 6.0447 & 6.0666 & 0.5242 & 1.7785 & 1.3332 & 2.5621 & 0.8421 & 3.5883 & 2.2340 & 5.0176 \\
            \hline\hline
            \multirow{3}{2.0cm}{Average Inference Time (sec)} & per cell & 0.0445 & 0.0461 & 0.0428 & 0.0421 &  0.0318 & 0.0294 & 0.0209 & 0.0182 & 0.0320 & 0.0358 & 0.0308 & 0.0290 \\
            \cline{2-14}
             & per object & 0.0900 & 0.0889 & 0.1097 & 0.1077 & 0.0769 & 0.0773 & 0.0801 & 0.0789 & 0.0595 & 0.0607 & 0.0896 & 0.0644 \\
            \cline{2-14}
             & per frame & 0.8961 & 0.9070 & 1.1151 & 1.0963 & 0.4834 & 0.4898 & 0.5067 & 0.5019 & 0.2543 & 0.2494 & 0.3525 & 0.2616 \\

             \hline\hline
             \multirow{3}{2.0cm}{Accuracy} & slice factor & \multicolumn{2}{c|}{3} & \multicolumn{2}{|c||}{1} & \multicolumn{2}{c|}{5} & \multicolumn{2}{|c||}{1} & \multicolumn{2}{c|}{5} & \multicolumn{2}{|c}{1} \\
             \cline{2-14}
              & RBDC & \multicolumn{2}{c|}{0.6018} & \multicolumn{2}{|c||}{0.5853} & \multicolumn{2}{c|}{0.3065} & \multicolumn{2}{|c||}{0.3001} & \multicolumn{2}{c|}{0.4540} & \multicolumn{2}{|c}{0.4421} \\
             \cline{2-14}
              & TBDC & \multicolumn{2}{c|}{0.7209} & \multicolumn{2}{|c||}{0.7117} & \multicolumn{2}{c|}{0.6603} & \multicolumn{2}{|c||}{0.6589} & \multicolumn{2}{c|}{0.8187} & \multicolumn{2}{|c}{0.8094} \\
            
            \end{tabular}
        \end{center}
        
    \end{table}
\end{landscape}

\section{Qualitative Results}
\subsection{Explainability}
\label{appendix:explainability}
In the following, we present explanations for example detections of anomalous events for all three benchmark datasets. The examples were chosen to cover the widest range of possible anomalies. For transparency reasons, we also cover an example in which our method failed to detect the correct anomaly in the video.

\subsubsection{CUHK Avenue}

In contrast to the example visualization in the main paper, we present a per cell explanation of the anomaly score for a sample image taken from the test video $03$ of CUHK Avenue. Here, a person is moving at an unusual speed through the scene which is considered anomalous in the given context. This example represents a motion-dependent anomaly.

\begin{figure}[!ht]
    \centering
    
    \begin{subfigure}{1.0\linewidth}
        \centering
        \includegraphics[width=\linewidth, center]{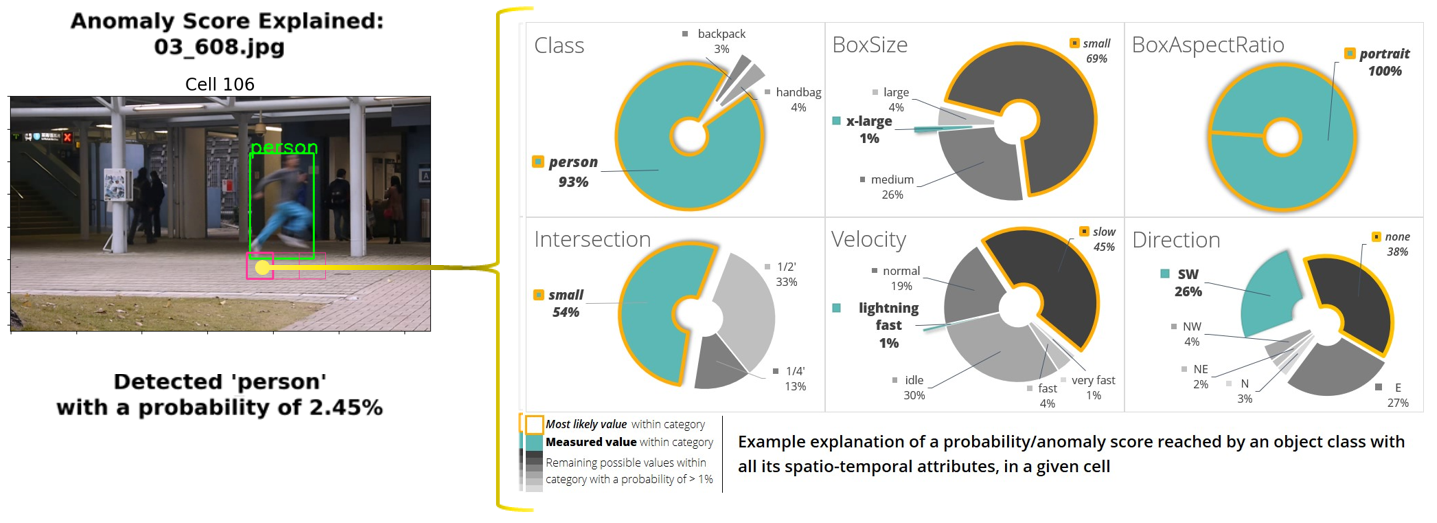}
        \caption{}
        \label{fig:106--explained}
    \end{subfigure}
    
    \begin{subfigure}{1.0\linewidth}
        \centering
        \includegraphics[width=\linewidth, center]{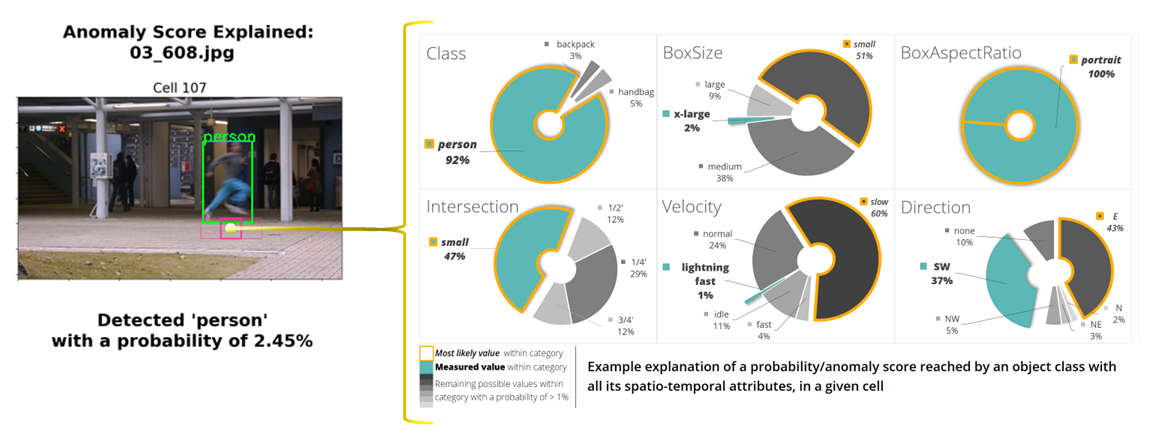}
        \caption{}
        \label{fig:107--explained}
    \end{subfigure}
    
    \begin{subfigure}{1.0\linewidth}
        \includegraphics[width=\linewidth, center]{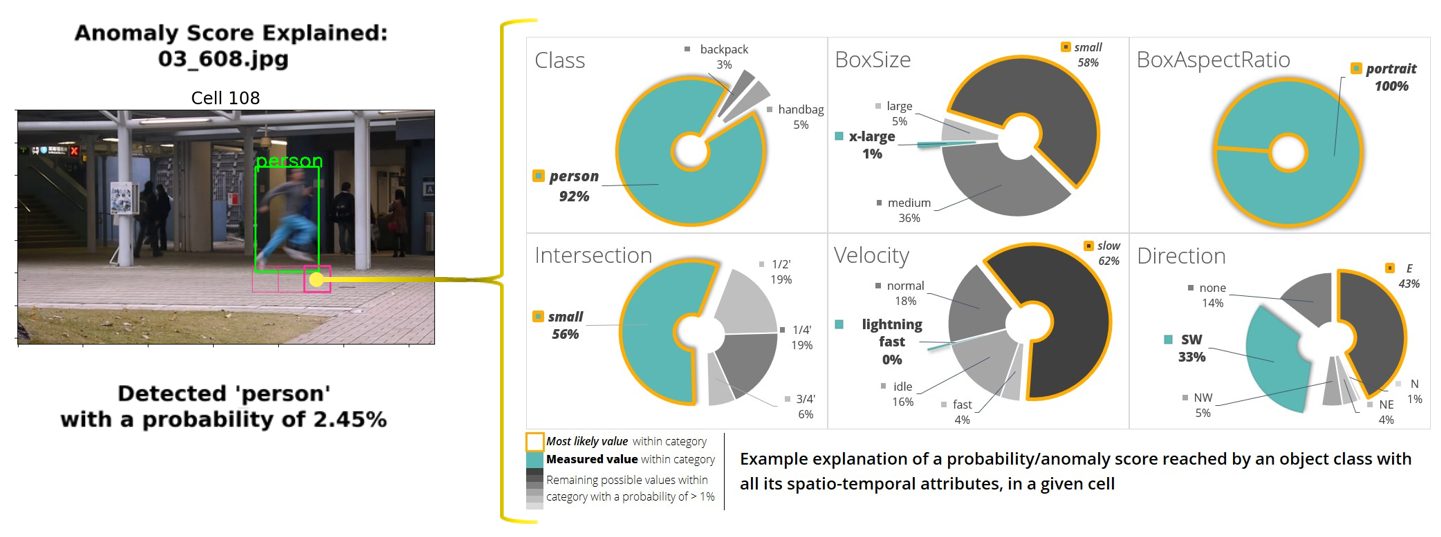}
        \caption{}
        \label{fig:108--explained}
    \end{subfigure}  
\caption{\textbf{Temporal Anomaly}: Explainability dashboards for test video $03$ of the benchmark dataset CUHK Avenue of all affected cells in which the given object appears.}
\label{fig:supplement--results-explained--avenue}
\end{figure}

From all three dashboards it can be deducted that apart from the velocity direction ("SW"), it is primarily the size ("x-large") and velocity ("lightning fast") of the given bounding box which are contributing most to the object's low probability. The reason for this is that first, the measured values within those two categories do not correspond to the most likely ones, i.e., the ones occupying the largest fraction of the respective chart, and that second, they in fact represent the least likely values among all others that were encountered during training. With the bottom border of the object's bounding box spanning three cells in total we thus present three explainability dashboards in~\Cref{fig:supplement--results-explained--avenue}.


\subsubsection{StreetScene}
In the example chosen from test video $31$ from StreetScene on the other hand, the present anomaly depicts a cyclist driving on the car lane of a street. Here, the anomaly is primarily related to its spatial location. Since under normal circumstances, i.e., during training, the most common class of objects detected in this area are vehicles (here, 'car' occupies $97\%$ of the chart area), the detected class 'person' contributes most to the low probability score of the object. The same can be concluded about the observed value of the RV \textit{Velocity}. In this case, the most probable velocity would be 'normal' while the actual detected value corresponds to 'very fast'. Similarly to the example in the main paper, here we present the average probability distributions over all affected cells.

\begin{figure}[!ht]
    \centering
    \includegraphics[width=1.0\textwidth, center]{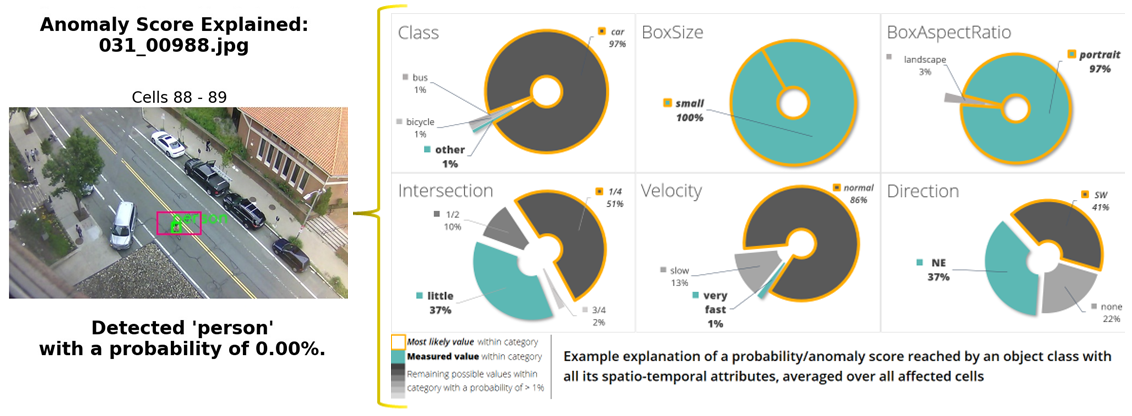}
    \caption{\textbf{Spatial Anomaly}: Explainability dashboard for test video $31$ of the benchmark dataset StreetScene of all affected cells in which the given object appears.\textbf{ A success case.}}
    \label{fig:results-explained--streetscene}
\end{figure}

\begin{figure}[!ht]
    \centering
    \includegraphics[width=1.0\textwidth, center]{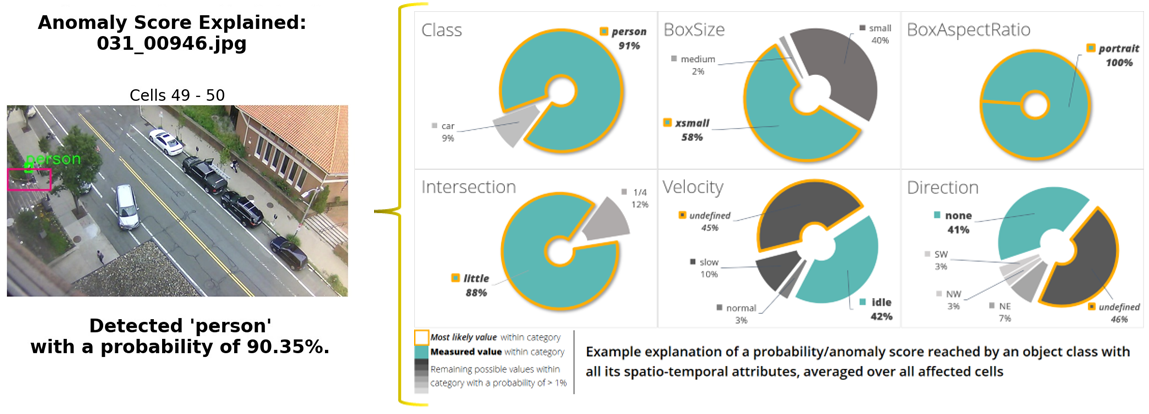}
    \caption{\textbf{Spatio-Temporal Anomaly}: Explainability dashboard for test video $31$ of the benchmark dataset StreetScene of all affected cells in which the given object appears.\textbf{ A failure case.}}
    \label{fig:results-explained--streetscene--fail}
\end{figure}

A few frames earlier in the same test video we chose to present the explainability charts for another object depicting a person sitting on the sidewalk which is also considered anomalous behavior. For transparency reasons, this particular example is meant to show in which cases our method fails to correctly identify anomalies, and shown in~\Cref{fig:results-explained--streetscene--fail}. As shown by the visualization, the probability of this particular object appearing in this location of the frame amounts to over $90\%$, and is thus very likely according to our discrete BN. The observed values of nearly all RVs always correspond to the most likely ones. The only two exceptions are object movement 'Velocity' and 'Direction'. In both cases, the observed values are equal to the second-most likely ones, i.e., 'idle' and 'none', respectively. The most likely ones in this particular case, 'undefined', denote edge scenarios in which the object was not found in the preceding frame, leaving both, movement velocity and direction undefined.

The most apparent reason for this false negative is the division of the frame into grid cells of a size that covers the entire width of the sidewalk. This implies that all pedestrians that pass by/shortly stop in this area during training will contribute significantly to making a person that sits there for a long time appear normal.

\subsection{More Results}
\begin{figure}[t!]
\begin{center}
    \begin{subfigure}{0.45\linewidth}
        \centering
        \includegraphics[width=1.0\textwidth, center]{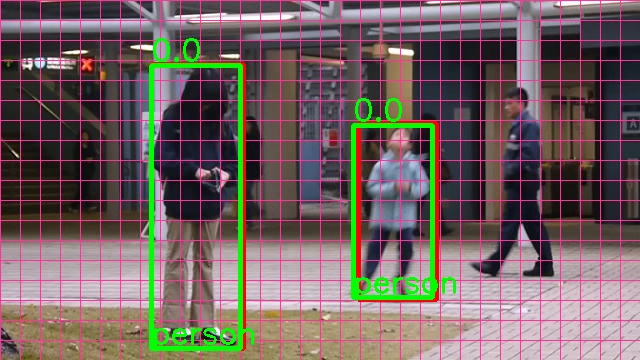}
        \caption{\textbf{CUHK Avenue} - Test video \#08}
    \end{subfigure}
    
    \begin{subfigure}{0.45\linewidth}
        \centering
        \includegraphics[width=1.0\textwidth, center]{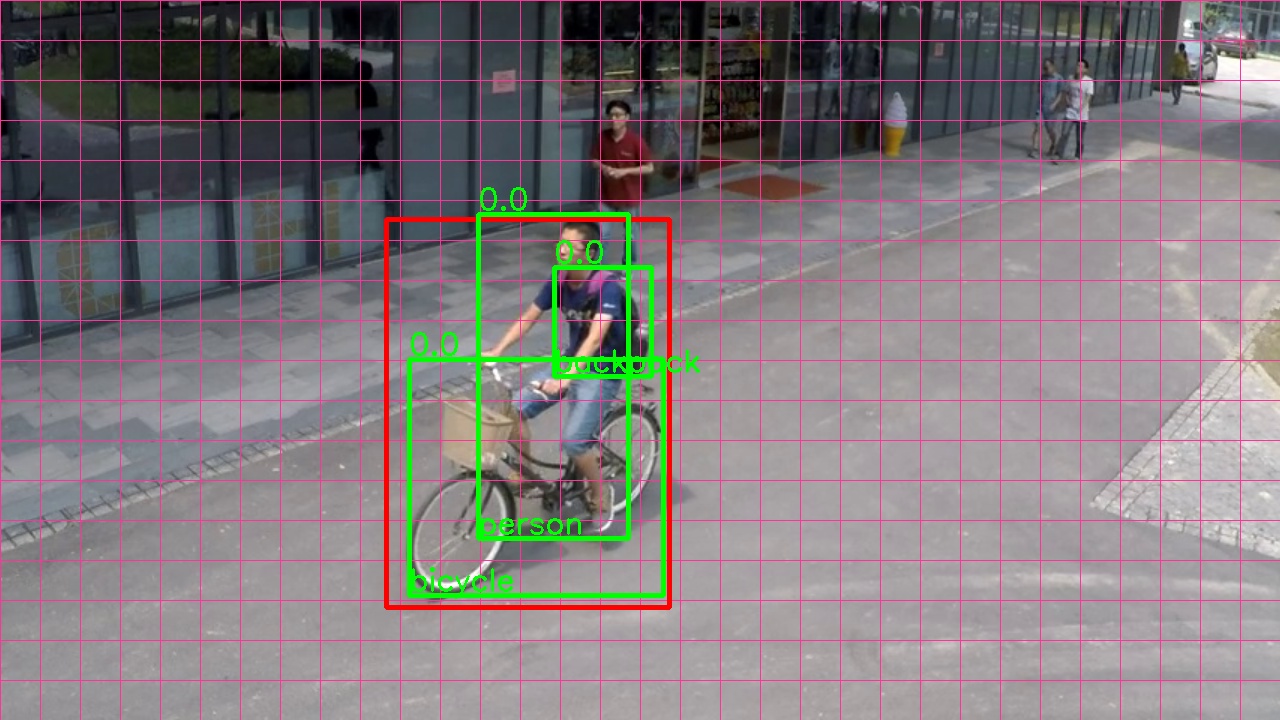}
        \caption{\textbf{ShanghaiTech} - Test video \#02}
    \end{subfigure}    
    \begin{subfigure}{0.45\linewidth}
        \centering
        \includegraphics[width=1.0\textwidth, center]{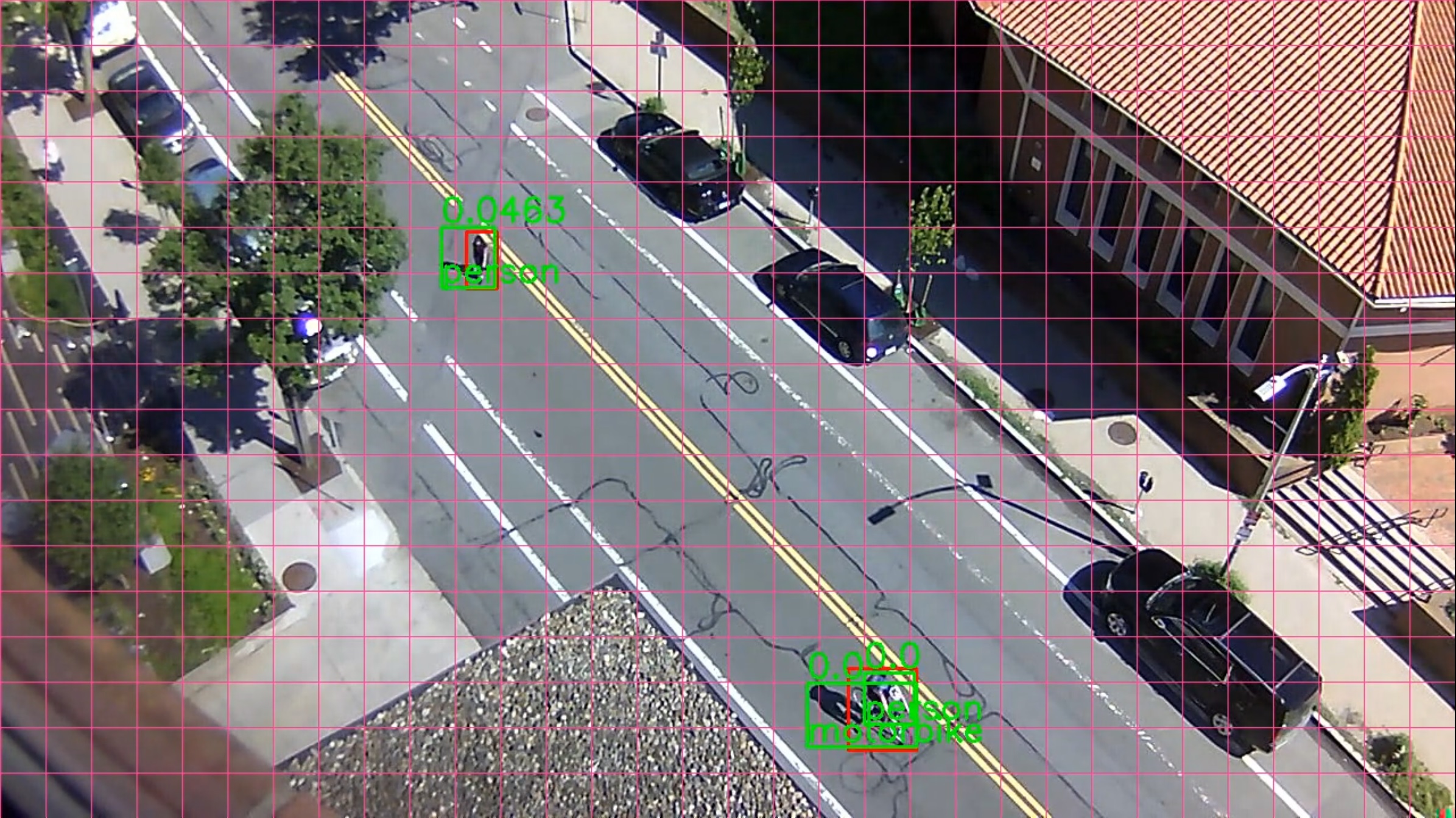}
        \caption{\textbf{StreetScene} - Test video \#16}
    \end{subfigure}
\end{center}
\caption{\textbf{Qualitative results} for test frames extracted from CUHK Avenue, ShanghaiTech and StreetScene. Ground Truth annotations are drawn in red, and detections in green (0 = anomalous, 1 = normal).}
\label{fig:visual-results--all-datasets}
\end{figure}

We present some additional visualizations of the final results yielded by our pipeline in~\Cref{fig:visual-results--all-datasets}. Shown are randomly chosen samples from all benchmark datasets.

\end{appendices}

\end{document}